\documentclass[journal]{IEEEtran}

\usepackage{cite}
\usepackage{amsmath,amssymb,amsfonts}
\usepackage{graphicx}
\usepackage{textcomp}
\usepackage{xcolor}
\usepackage{tikz,pgfplots}
\usetikzlibrary{shapes,shapes.geometric}
\usepackage{array}
\usepackage{booktabs}
\usepackage{multirow}
\usepackage{makecell}
\usepackage{microtype}
\usepackage{url}
\usepackage[hidelinks]{hyperref}
\pgfplotsset{compat=1.18}

\begin{document}

\title{Mechanical Conscience: A Mathematical Framework for Dependability of Machine Intelligence}

\author{Munkhdegerekh~Batzorig,
        Purevbaatar~Ganbold,
        Kyungbin~Park,
        Pilkong~Jeong,
        and~Kangbin~Yim%
\thanks{M. Batzorig, P. Ganbold, K. Park, and K. Yim are with the Department of Information Security Engineering, Soonchunhyang University, Asan, South Korea (e-mail: munkhdelgerekh@sch.ac.kr; puujee12@sch.ac.kr; xeraph@sch.ac.kr; yim@sch.ac.kr).}%
\thanks{P. Jeong is with the Department of Mobility Convergence Security, Soonchunhyang University, Asan, South Korea (e-mail: pkjeong@sch.ac.kr).}}

\maketitle

\begin{abstract}
Distributed collaborative intelligence (DCI)---encompassing
edge-to-edge architectures, federated learning, transfer learning, and swarm
systems---creates environments in which emergent risk is structurally
unavoidable: locally correct decisions by individual agents compose into
globally unacceptable behavioral trajectories under uncertainty. Existing
approaches such as constrained optimization, safe reinforcement learning, and
runtime assurance evaluate acceptability at the level of individual actions
rather than across behavioral trajectories, and none addresses the
multi-participant, uncertainty-laden nature of DCI deployments. This paper
introduces \emph{mechanical conscience} (MC), a novel concept and simplified
mathematical framework that operationalizes trajectory-level normative
regulation for both single-agent and distributed intelligent systems.
Mechanical conscience is defined as a supervisory filter that minimally
corrects a baseline policy's actions to reduce cumulative deviation from a
normatively admissible region, while accounting for epistemic uncertainty.
We introduce associated constructs---\emph{conscience score},
\emph{mechanical guilt}, and \emph{resonant dependability}---that provide an
interpretable vocabulary and computable governance signals for this emerging
field. Core theoretical properties are established: admissibility equivalence,
existence of optimal regulation, and monotonic deviation reduction.
Illustrative results demonstrate that MC-regulated agents maintain
trajectory-level normative acceptability where conventional controllers drift
outside admissible bounds, and that the framework naturally extends to
suppress interaction-induced emergent risk in multi-agent DCI settings.
\end{abstract}


\section{Introduction}
\label{sec:introduction}

\IEEEPARstart{T}{he} deployment of machine intelligence has moved well beyond isolated,
centralized computing. Contemporary systems increasingly operate as
\emph{distributed collaborative intelligence} (DCI)---a broad family of
architectures in which autonomous or semi-autonomous agents learn, reason,
and act in a shared environment without any single point of global
coordination.

Prominent DCI paradigms illustrate the breadth of this shift.
\emph{Edge-to-edge (E2E) architectures}~\cite{Shi2016,Satyanarayanan2017}
push inference and decision-making to the network edge, enabling real-time
physical AI in autonomous vehicles and industrial automation, but at the cost
of fragmented situational awareness across the system.
\emph{Federated learning}~\cite{McMahan2017,Li2020FL} coordinates model
training across heterogeneous participants without centralizing raw data,
making privacy preservation possible while rendering normative divergence
across participants nearly inevitable.
\emph{Transfer learning}~\cite{Pan2010} allows agents trained in one domain
to be rapidly deployed in another, amplifying capability while also
transplanting behaviors that may be contextually inappropriate or normatively
misaligned in the target environment.
\emph{Swarm and swarm-inspired learning}~\cite{Dorigo2021} coordinates large
populations of simple agents through local interaction rules, producing
collective intelligence whose global behavior cannot be predicted from any
individual agent's local policy.

What these paradigms share is a structural property that makes safety analysis
fundamentally harder: \emph{emergent risk}. In a DCI environment, each agent
may behave correctly with respect to its own local objectives and constraints,
yet the composition of many such locally correct decisions---across time and
across interacting agents---can produce globally unacceptable outcomes.
Emergent risk is not caused by individual failures; it arises from the
trajectory-level accumulation of small normative deviations under uncertainty.

Current technical solutions address individual facets of this problem but do
not resolve it. Constrained optimal control enforces instantaneous feasibility
but ignores how trajectories drift over time. Safe reinforcement learning
imposes policy-level constraints but does not prevent uncertainty-induced norm
violations at runtime. Runtime assurance frameworks react to breaches but apply
binary interventions rather than continuous, graduated regulation. None of
these approaches provides a \emph{trajectory-level} model of normative
compliance that operates continuously, accounts for uncertainty, and extends
naturally to the distributed multi-agent settings characteristic of DCI.

This paper introduces \textbf{mechanical conscience} (MC)---a new concept
and formal mathematical framework that addresses this gap. The name
intentionally draws an analogy to human conscience: not a rigid rule system,
but a continuously active internal regulatory mechanism that evaluates whether
one's behavior remains within acceptable normative bounds across time. In DCI
environments, mechanical conscience functions as a shared regulatory substrate
at the trajectory level, enabling what we term \emph{resonant dependability}:
the emergent, bidirectional trust between human and machine intelligence that
arises from sustained normative alignment.

The key contributions of this paper are as follows:
\begin{itemize}
  \item We formally define \emph{mechanical conscience} as a trajectory-level
        normative supervisory functional, with associated constructs:
        conscience score, conscience deviation, and mechanical guilt.
  \item We introduce \emph{resonant dependability} as the emergent trust that
        arises when human and machine intelligence sustain mutually acknowledged
        normative trajectories over time.
  \item We establish core theoretical properties: admissibility equivalence,
        existence of optimal regulation, and monotonic deviation reduction.
  \item We extend the formulation to distributed DCI settings, capturing
        emergent risk at individual, pairwise, and collective levels.
\end{itemize}

\section{Related Work}
\label{sec:related_work}

The mechanical conscience framework intersects several distinct research
threads: distributed collaborative intelligence and its emergent risks,
constrained and shielded reinforcement learning, ethical-governor and
consequence-engine architectures, deontic and normative multi-agent
reasoning, AI alignment, and classical dependability theory. This section
surveys each thread, identifies its limits relative to trajectory-level
normative regulation in DCI, and positions mechanical conscience within
the resulting landscape.

\subsection{Distributed Collaborative Intelligence and Emergent Risk}

The emergence of distributed, multi-participant intelligence architectures
has introduced normative challenges that classical system safety frameworks
were not designed to address.

\emph{Edge-to-edge (E2E) computing}~\cite{Shi2016,Satyanarayanan2017}
relocates inference from centralized clouds to network-edge devices,
enabling latency-sensitive applications such as autonomous driving and
robotic surgery. However, E2E architectures fragment global situational
awareness: no single node observes the full system state, and decisions
made locally and independently can compose into globally unacceptable
trajectories.

\emph{Federated learning}~\cite{McMahan2017,Li2020FL} allows many
participants to jointly train a shared model without revealing their raw
data. While this design preserves privacy, it also decentralizes both
learning and policy formation. Different participants may optimize under
different implicit norms, and the aggregated model can inherit or amplify
normative inconsistencies that are invisible to any individual participant.
Recent surveys of \emph{trustworthy federated
learning}~\cite{TrustworthyFL2024,TrustworthyDistAI2024} catalog these
challenges along the axes of robustness, fairness, and privacy, but treat
each as a static property of the trained model rather than as a property
of behavioral trajectories evolving over time. The mechanical conscience
framework complements this body of work by providing a runtime,
trajectory-level normative regulator that operates after deployment and is
agnostic to how the underlying federated model was trained.

\emph{Transfer learning}~\cite{Pan2010} accelerates deployment by
initializing agents with knowledge from a source domain. Yet behaviors
that were normatively appropriate in the source domain may be misaligned
in the target environment. When such transferred agents are deployed in
interactive DCI settings, their normative misalignment can propagate to
other agents through shared observations or joint action spaces.

\emph{Swarm and swarm-inspired learning}~\cite{Dorigo2021} coordinates
agents through local interaction rules, yielding scalable collective
behavior. The global outcomes of swarm coordination are, by design,
emergent---they are not specified by any individual agent's policy. This
makes swarm systems particularly susceptible to interaction-induced
normative drift, where an accumulation of locally innocuous decisions
produces globally unacceptable collective behavior.

Across all of these DCI paradigms, a common pattern emerges: locally
rational behavior composes into globally problematic trajectories under
uncertainty. The mechanical conscience framework is designed precisely to
address this structural gap.

\subsection{Constrained Control and Safety Filters}

\emph{Constrained optimal control} and \emph{control barrier functions}
(CBFs)~\cite{Ames2019,Boyd2004,Prajna2004} enforce instantaneous safety or
feasibility conditions on state-action pairs. CBF-based methods provide
strong forward-invariance guarantees for a predefined safe set, but they
evaluate acceptability at each time step independently and do not model
how cumulative trajectory drift can render a system normatively
unacceptable over time. Mechanical conscience complements these approaches
by providing the \emph{trajectory-level} regulation that point-wise
methods inherently lack, while reusing them as natural building blocks
for the satisfaction functions $\phi_i$ in Section~\ref{sec:formulation}.

\subsection{Safe Reinforcement Learning and Shielding}

\emph{Safe reinforcement learning}~\cite{Garcia2015,Altman1999,Berkenkamp2017}
incorporates safety constraints into the policy learning objective,
typically through constrained Markov decision processes or Lyapunov-based
methods. These approaches constrain policy behavior in expectation over
the training distribution and do not directly prevent transient or
uncertainty-induced norm violations at deployment time.

A particularly relevant subfield is \emph{shielded reinforcement
learning}~\cite{Alshiekh2018}, in which an external runtime filter---the
shield---restricts agent actions or corrects undesired ones to enforce
formal safety specifications, typically expressed in temporal logic.
Shielding has since been extended to dynamic temporal-logic
constraints~\cite{ShieldedSTL2026} and to multi-agent settings with
\emph{model-based dynamic shielding} (MBDS)~\cite{Xiao2023MBDS}, which
synthesizes shields online to mitigate the conservativeness of static
multi-agent shields. Mechanical conscience shares the supervisory-filter
philosophy of shielding but differs along three axes that are essential
for DCI deployment:
\begin{enumerate}
  \item Shields enforce \emph{hard binary constraints} derived from formal
        specifications; mechanical conscience produces \emph{continuous,
        graduated} corrections via the deviation functional $\Psi$,
        accommodating soft norms (fairness, privacy, social acceptability)
        that resist hard logical specification.
  \item Shields require formal synthesis from specification and a model
        of the environment; mechanical conscience requires only
        differentiable satisfaction functions and is therefore architecture-
        and policy-agnostic.
  \item Shields are typically defined for a single normative dimension
        (safety); mechanical conscience aggregates a heterogeneous
        normative space $\mathcal{N}$ that natively spans safety,
        security, fairness, regulatory compliance, and other domains.
\end{enumerate}
Recent safe-MARL methods such as MADAC~\cite{Li2024SafeMARL}, which
enforces state-wise constraints with feasibility guarantees, and
entropic-exploration coordination~\cite{E2C2024}, which treats safety as
a team-level concept, are directly relevant to the multi-agent extension
in Section~\ref{sec:dependability} and are natural candidates for
empirical comparison in future work.

\subsection{Ethical Governor and Consequence-Engine Architectures}

The closest conceptual antecedents of mechanical conscience are the
\emph{ethical governor} and \emph{consequence-engine} architectures
developed in the machine ethics community.

Arkin's ethical governor~\cite{Arkin2009} introduced an architectural
component that suppresses, restricts, or transforms the lethal output of
an autonomous system so that it falls within a permissible behavior set
defined by laws of war and rules of engagement. Winfield et
al.~\cite{Winfield2014} proposed an internal-model-based
\emph{consequence engine} that simulates the predicted outcomes of
candidate actions and filters them through a safety/ethical logic.
Dennis et al.~\cite{Dennis2015} subsequently provided formal verification
techniques for the ethical decision-making of such governors. Most
recently, neuro-symbolic ethical governors~\cite{NeuroSymbolicEG2026} and
the GRACE architecture~\cite{GRACE2026} integrate learning-based
risk-assessment components with symbolic obligation reasoning, separating
normative reasoning from instrumental optimization to enable
interpretability and contestability.

Mechanical conscience extends this lineage in three substantive ways.
First, governor and consequence-engine architectures evaluate
acceptability of \emph{individual actions} or short-horizon plans;
mechanical conscience operates over \emph{behavioral trajectories} and
introduces an explicit cumulative deviation signal (mechanical guilt
$G_T$) that is auditable in real time. Second, existing governors are
typically defined for single agents or small teams; mechanical conscience
is formulated from the outset for distributed multi-agent DCI through
the additive decomposition $\Psi_{\mathrm{tot}} = \sum_k \Psi^{(k)} +
\sum_{i<j}\Psi_{ij}+\Psi_{\mathrm{global}}$, which captures
interaction-induced emergent risk explicitly. Third, governors and
consequence engines focus on the supervisory mechanism itself; mechanical
conscience pairs that mechanism with the systems-level concept of
\emph{resonant dependability}, articulating the long-term trust property
that the supervisory layer is intended to produce.

\subsection{Deontic Logic and Normative Multi-Agent Systems}

Deontic logic and normative multi-agent
systems~\cite{Boella2008NorMAS,Furbach2015} provide a formal foundation
for reasoning about obligations, permissions, and prohibitions, with
recent extensions including defeasible deontic
calculi~\cite{Olson2024DDIC} for resolving norm conflicts and deontic
temporal logics~\cite{DeonticTemporal2026} for formal verification of AI
ethical properties. These approaches are highly expressive but operate
at the symbolic level and produce decisions of the form
``permitted/forbidden/obligatory.'' Mechanical conscience can be viewed
as a \emph{quantitative companion} to deontic reasoning: each norm
$N_i \in \mathcal{N}$ may be derived from a deontic specification, but
the satisfaction function $\phi_i$ converts that specification into a
real-valued degree of compliance, enabling continuous regulation through
$\Psi$ and integration into gradient-based control loops. This
quantitative bridge between symbolic norms and continuous control is, to
our knowledge, not provided by existing deontic frameworks.

\subsection{AI Alignment and Value Specification}

\emph{AI alignment}~\cite{Russell2019,Amodei2016} research investigates
how to ensure that AI systems pursue objectives consistent with human
values. Contemporary alignment methods range from reward modeling and
inverse reinforcement learning to constitutional AI and reinforcement
learning from human feedback. While these methods address \emph{what}
objectives a system should hold, they do not provide a runtime
operational model for \emph{how} those objectives are continuously
enforced at the trajectory level under uncertainty and in interaction
with other agents. Mechanical conscience is intended to serve as
precisely this operational layer: the mechanism by which alignment
objectives, once specified, are enacted in real time within a DCI
environment.

\subsection{Dependability and Trust in Distributed Systems}

Classical dependability theory~\cite{Laprie1992} characterizes system
trustworthiness through attributes such as reliability, availability, and
safety. These attributes are defined for systems in isolation and are
evaluated against static specifications. They do not capture the
\emph{relational} and \emph{trajectory-aware} nature of trust in
human--machine DCI settings, where dependability is not a static property
of one system but an emergent property of sustained interaction between
human and machine intelligence. The concept of \emph{resonant
dependability} introduced in this paper extends classical dependability
theory to this relational, trajectory-level setting, with mechanical
conscience as its operational realization.

\subsection{Positioning of Mechanical Conscience}

Table~\ref{tab:related_comparison} summarizes the positioning of
mechanical conscience relative to the principal threads surveyed above.
Mechanical conscience is distinguished by the simultaneous combination
of (i) trajectory-level rather than point-wise evaluation, (ii) a
continuous, differentiable, multi-norm deviation functional, (iii) a
native multi-agent decomposition capturing interaction-induced emergent
risk, and (iv) integration with the systems-level concept of resonant
dependability.

\begin{table*}[t]
\centering
\caption{Positioning of mechanical conscience (MC) relative to related threads.
\checkmark: addressed; --: not addressed or out of scope; $\triangle$: partially addressed.}
\label{tab:related_comparison}
\renewcommand{\arraystretch}{1.2}
\setlength{\tabcolsep}{8pt}
\footnotesize
\begin{tabular}{@{}lcccc@{}}
\toprule
\textbf{Approach} & \textbf{Trajectory-level} & \textbf{Continuous} & \textbf{Multi-norm} & \textbf{Multi-agent} \\
                  & \textbf{evaluation}       & \textbf{regulation} & \textbf{aggregation} & \textbf{native}     \\
\midrule
Constrained / CBF control             & --          & \checkmark  & --          & $\triangle$ \\
Safe RL (CMDP)                        & $\triangle$ & --          & --          & --           \\
Shielding (LTL/STL)                   & \checkmark  & --          & --          & $\triangle$  \\
Ethical governor / consequence engine  & $\triangle$ & $\triangle$ & $\triangle$ & --           \\
Deontic / normative MAS               & $\triangle$ & --          & \checkmark  & \checkmark   \\
AI alignment                          & $\triangle$ & --          & $\triangle$ & --           \\
\textbf{Mechanical conscience (this work)} & \checkmark & \checkmark & \checkmark & \checkmark  \\
\bottomrule
\end{tabular}
\end{table*}

\section{System and Normative Formulation}
\label{sec:formulation}

\subsection{Dynamical System Model}

We model an intelligent agent---or a DCI system abstracted as a single
decision-making entity---as a discrete-time stochastic dynamical system:
\begin{equation}
  x_{t+1} = f(x_t,\, u_t,\, w_t), \label{eq:dynamics}
\end{equation}
where $x_t \in \mathcal{X}$ is the system state at time $t$,
$u_t \in \mathcal{U}$ is the control action, and $w_t \in \mathcal{W}$
captures environmental disturbances and model uncertainty.
The system produces observations
\begin{equation}
  y_t = h(x_t, v_t),
\end{equation}
where $v_t$ is sensing noise. A baseline policy $\pi_\theta$ maps the
estimated state $\hat{x}_t$ to a candidate action:
\begin{equation}
  u_t^{\mathrm{base}} = \pi_\theta(\hat{x}_t).
\end{equation}
The baseline policy may be a learned policy (e.g., from federated or
transfer learning), a hand-crafted controller, or any combination. Mechanical
conscience operates as a post-hoc supervisory layer that does not require
access to the internals of $\pi_\theta$.

\subsection{Normative Evaluation Space}

We introduce a \emph{normative evaluation space}
$\mathcal{N} = \{N_1, N_2, \ldots, N_m\}$,
where each $N_i$ represents a normative category relevant to the deployment
context---safety, security, privacy, fairness, mission compliance, or
regulatory requirement. Each norm induces a real-valued \emph{satisfaction
function}:
\begin{equation}
  \phi_i(x_t, u_t, \xi_t) \in \mathbb{R},
\end{equation}
where $\xi_t$ encodes contextual variables such as environmental state,
social context, or applicable regulations.
The sign convention is simple: $\phi_i \geq 0$ means norm $N_i$ is
satisfied; $\phi_i < 0$ means it is violated with severity $|\phi_i|$.

The \textbf{conscience-feasible set}---the normative admissible region---at
time $t$ is:
\begin{equation}
  \mathcal{C}_t = \bigl\{(x_t, u_t) : \phi_i(x_t, u_t, \xi_t) \geq 0,\; \forall i\bigr\}.
  \label{eq:feasible}
\end{equation}
This set captures not merely what is operationally possible, but what remains
\emph{acceptable under the active normative regime}.

\subsection{Conscience Score and Normative Deviation}

A scalar \textbf{conscience score} summarizing overall normative acceptability
is:
\begin{equation}
  \Gamma_t = \sum_{i=1}^{m} \alpha_i\, \phi_i(x_t, u_t, \xi_t),
  \label{eq:conscience_score}
\end{equation}
where $\alpha_i \geq 0$ is the importance weight of norm $N_i$.
High $\Gamma_t$ indicates a morally or operationally acceptable state-action
pair; low $\Gamma_t$ signals a conscience alarm.

The \textbf{normative deviation functional} isolates the cost of violations:
\begin{equation}
  \Psi(x_t, u_t) = \sum_{i=1}^{m} \alpha_i \max\bigl(0,\, -\phi_i(x_t, u_t, \xi_t)\bigr),
  \label{eq:deviation}
\end{equation}
which is zero if and only if $(x_t, u_t) \in \mathcal{C}_t$, and equals the
weighted sum of violation magnitudes otherwise.
The linear form (exponent $p=1$) is intentionally simple: it is
differentiable almost everywhere, convex, and interpretable as a
violation cost that scales proportionally with severity.

\section{The Mechanical Conscience Framework}
\label{sec:framework}

\subsection{Mechanical Conscience as a Supervisory Filter}

Mechanical conscience is realized as a \emph{supervisory filter} sitting
between the baseline policy and the actuators. Given a candidate action
$u_t^{\mathrm{base}} = \pi_\theta(\hat{x}_t)$ produced by the baseline
policy, the MC layer computes the minimally corrected action
\begin{equation}
  u_t^* = \arg\min_{u \in \mathcal{U}}
  \Bigl[\,\|u - u_t^{\mathrm{base}}\|^2 \;+\; \eta\,\Psi(\hat{x}_t, u)\,\Bigr],
  \label{eq:supervisory}
\end{equation}
where $\eta > 0$ controls the strength of normative correction.
The quadratic term $\|u - u_t^{\mathrm{base}}\|^2$ enforces \emph{minimal
intervention}: the MC layer deviates from the original intent only as much
as is necessary to restore normative acceptability. This is the operational
core of mechanical conscience---a lightweight, architecture-neutral
post-processing step that can be inserted after any baseline policy without
modifying the policy itself. The formulation is deliberately simple:
$\Psi$ provides a smooth scalar penalty, the quadratic deviation term is
strictly convex, and the resulting optimization can be solved efficiently
by standard quadratic programming or gradient-based methods. When all
$\phi_i$ are linear in $u$, the problem reduces to a quadratic program
with linear constraints; when $\phi_i$ are nonlinear, projected gradient
descent suffices in practice.

\subsection{Trajectory Conscience Functional}

A point-in-time conscience score is insufficient; true conscience evaluates
the \emph{projected consequences} of current actions over a future horizon.
We define the \textbf{mechanical conscience functional} as the expected
cumulative trade-off between task utility and normative cost:
\begin{equation}
  \mathcal{J}^{\mathrm{MC}} = \mathbb{E}\!\left[
    \sum_{k=0}^{H} \gamma^k
    \Bigl(R(x_{t+k}, u_{t+k}) - \beta\,\Psi(x_{t+k}, u_{t+k})\Bigr)
  \right],
  \label{eq:mc_functional}
\end{equation}
where $R(x_t, u_t) \geq 0$ is the operational task reward, $\Psi$ is the
normative deviation functional (Eq.~\ref{eq:deviation}), $\beta > 0$ is the
\emph{conscience strength} that governs the normative-utility trade-off,
$\gamma \in (0,1]$ is a temporal discount factor, and $H$ is the planning
horizon. Maximizing $\mathcal{J}^{\mathrm{MC}}$ yields trajectories that
prefer useful actions but penalize downstream normative deviation.
The supervisory filter~\eqref{eq:supervisory} is the per-step instantiation
of this objective: with $H = 0$ and $\eta = \beta$, minimizing the negative
of $\mathcal{J}^{\mathrm{MC}}$ plus the minimal-intervention regularizer
recovers~\eqref{eq:supervisory}. For $H > 0$, mechanical conscience
becomes a horizon-aware regulator naturally compatible with model
predictive control architectures.

\subsection{Uncertainty-Aware Conscience}

A realistic conscience must account for epistemic uncertainty.
In DCI environments, normative classification is often uncertain:
the applicable norm, the context $\xi_t$, or the state estimate
$\hat{x}_t$ may be unreliable. Let $p(x_t \mid y_{1:t})$ be the state
belief distribution maintained by the agent, and let
$p(c \mid \hat{x}_t, u_t)$ denote a posterior over normative classes
$c$ (e.g., admissible / inadmissible) produced by a normative classifier
operating on the estimated state-action pair. We define the
\emph{uncertainty severity} as
\begin{equation}
  \Omega_t = 1 - \max_c\, p(c \mid \hat{x}_t, u_t),
  \label{eq:uncertainty}
\end{equation}
which lies in $[0,1]$ and is high when the system cannot confidently
classify the current state-action pair as normatively acceptable.
An equivalent variance-based form
$\Omega_t = \mathrm{Var}\!\left[\Psi(x_t,u_t)\right]$
is appropriate when $\Psi$ is treated as a random functional under the
state belief.

The full \textbf{uncertainty-augmented conscience penalty} is then
\begin{equation}
  \Psi^{\mathrm{uc}}(x_t, u_t) = \Psi(x_t, u_t) + \rho\,\Omega_t,
  \label{eq:uc_penalty}
\end{equation}
with $\rho \geq 0$ controlling the degree of uncertainty aversion.
Substituting $\Psi^{\mathrm{uc}}$ for $\Psi$ in
either~\eqref{eq:supervisory} or~\eqref{eq:mc_functional} yields the
uncertainty-aware variants of the supervisory filter and the trajectory
conscience functional. This substitution ensures the system becomes
\emph{more conservative} when it is uncertain---a property consistent
with how human conscience operates under moral ambiguity, and
particularly important in DCI deployments where uncertainty is
heterogeneous across participants and changes dynamically.

\subsection{Conscience Deviation and Mechanical Guilt}

To support interpretability and runtime monitoring, we introduce two
cumulative quantities that summarize the normative behavior of the
system over time. The \textbf{conscience deviation} at time $t$ is the
instantaneous normative violation cost,
$D_t = \Psi(x_t, u_t)$, equivalently $\Psi^{\mathrm{uc}}(x_t,u_t)$ when
the uncertainty-aware form is used.
Its discounted accumulation over a horizon $T$---termed
\textbf{mechanical guilt}---is
\begin{equation}
  G_T = \sum_{t=0}^{T} \gamma^t\, D_t.
  \label{eq:guilt}
\end{equation}
A system with healthy mechanical conscience minimizes $G_T$, not merely
maximizing raw task reward. Mechanical guilt is computable in real time,
requires no additional model beyond $\Psi$, and is directly auditable by
human overseers, making accumulated normative stress a practical
governance signal for DCI deployments. The relationship between
$\mathcal{J}^{\mathrm{MC}}$ and $G_T$ is direct: under any
$\beta$-regulated policy, $\mathbb{E}[G_T]$ is the running cost
component of $-\mathcal{J}^{\mathrm{MC}}$, so improving the conscience
functional is equivalent to reducing expected mechanical guilt while
preserving task utility.

\section{Theoretical Properties}
\label{sec:properties}

We establish the foundational mathematical properties of the mechanical
conscience framework. The three principal propositions characterize
\emph{faithfulness}, \emph{well-posedness}, and \emph{tunability} of MC
regulation in the single-agent setting; we then state corollaries that
extend these properties to the uncertainty-aware variant
($\Psi^{\mathrm{uc}}$, Eq.~\ref{eq:uc_penalty}) and to the distributed
multi-agent setting (Section~\ref{sec:dependability}).
All proofs follow directly from the definitions in
Sections~\ref{sec:formulation} and~\ref{sec:framework}.

\medskip
\noindent\textbf{Proposition 1 (Admissibility Equivalence).}
\textit{The normative deviation functional satisfies}
\begin{equation}
  \Psi(x_t, u_t) = 0 \iff (x_t, u_t) \in \mathcal{C}_t.
\end{equation}

\noindent\textit{Proof.}
By Eq.~\ref{eq:deviation},
$\Psi(x_t, u_t) = \sum_{i=1}^{m} \alpha_i \max\bigl(0, -\phi_i(x_t,u_t,\xi_t)\bigr)$.
Since $\alpha_i > 0$ for all $i$ and each summand is non-negative, $\Psi=0$
holds if and only if $\max\bigl(0, -\phi_i\bigr)=0$ for every $i$, which
is equivalent to $\phi_i(x_t,u_t,\xi_t) \geq 0$ for all $i$. This is
precisely the defining condition of $\mathcal{C}_t$ in
Eq.~\ref{eq:feasible}.
\hfill$\square$

\medskip
\noindent\textbf{Proposition 2 (Existence of Optimal Regulation).}
\textit{If $\mathcal{U}$ is compact and each $\phi_i$ is continuous in
$u$, then the supervisory optimization
problem~\eqref{eq:supervisory} admits a minimizer $u_t^* \in \mathcal{U}$.}

\noindent\textit{Proof.}
Since each $\phi_i$ is continuous in $u$ and the maps
$z \mapsto \max(0,-z)$ and finite addition preserve continuity,
$\Psi(\hat{x}_t, \cdot)$ is continuous on $\mathcal{U}$. The Euclidean norm
$\|u - u_t^{\mathrm{base}}\|^2$ is also continuous in $u$. The objective
in~\eqref{eq:supervisory} is therefore a continuous function on the
compact set $\mathcal{U}$, and by the extreme value theorem it attains
its infimum at some $u_t^* \in \mathcal{U}$.
\hfill$\square$

\noindent\textit{Remark.}
The continuity assumption on $\phi_i$ is mild: it is satisfied by all
common normative constructions used in practice, including barrier-
function-style constraints~\cite{Ames2019}, smooth distance-based norms,
and bounded outputs of differentiable normative classifiers. When all
$\phi_i$ are linear in $u$, the optimization in~\eqref{eq:supervisory}
becomes a strictly convex quadratic program with a unique solution.

\medskip
\noindent\textbf{Proposition 3 (Monotonic Regulation).}
\textit{Let $\pi_\beta^*$ denote the optimal policy maximizing the
mechanical conscience functional~\eqref{eq:mc_functional} for a given
conscience strength $\beta \geq 0$, and let $G_T^{\beta}$ denote the
expected mechanical guilt accumulated under $\pi_\beta^*$ over horizon
$T$. Then $\beta \mapsto \mathbb{E}[G_T^{\beta}]$ is non-increasing.}

\noindent\textit{Proof.}
Fix $0 \leq \beta_1 \leq \beta_2$. By optimality of $\pi_{\beta_1}^*$
under $\mathcal{J}^{\mathrm{MC}}$ with conscience strength $\beta_1$,
\begin{figure*}[t]
\normalsize
\begin{equation}
  \mathbb{E}_{\pi_{\beta_1}^*}\!\left[\sum_{k=0}^{H} \gamma^k R_k\right]
  - \beta_1 \mathbb{E}_{\pi_{\beta_1}^*}\!\left[\sum_{k=0}^{H} \gamma^k \Psi_k\right]
  \;\geq\;
  \mathbb{E}_{\pi_{\beta_2}^*}\!\left[\sum_{k=0}^{H} \gamma^k R_k\right]
  - \beta_1 \mathbb{E}_{\pi_{\beta_2}^*}\!\left[\sum_{k=0}^{H} \gamma^k \Psi_k\right],
  \label{eq:opt_beta1}
\end{equation}
where for brevity $R_k = R(x_{t+k},u_{t+k})$ and
$\Psi_k = \Psi(x_{t+k},u_{t+k})$ under the indicated policy.
Symmetrically, by optimality of $\pi_{\beta_2}^*$ under conscience
strength $\beta_2$,
\begin{equation}
  \mathbb{E}_{\pi_{\beta_2}^*}\!\left[\sum_k \gamma^k R_k\right]
  - \beta_2 \mathbb{E}_{\pi_{\beta_2}^*}\!\left[\sum_k \gamma^k \Psi_k\right]
  \;\geq\;
  \mathbb{E}_{\pi_{\beta_1}^*}\!\left[\sum_k \gamma^k R_k\right]
  - \beta_2 \mathbb{E}_{\pi_{\beta_1}^*}\!\left[\sum_k \gamma^k \Psi_k\right].
  \label{eq:opt_beta2}
\end{equation}
\hrulefill
\end{figure*}
Adding~\eqref{eq:opt_beta1} and~\eqref{eq:opt_beta2} and simplifying yields
\begin{equation}
  (\beta_2 - \beta_1) \left(
    \mathbb{E}_{\pi_{\beta_1}^*}\!\left[\sum_k \gamma^k \Psi_k\right]
    \;-\;
    \mathbb{E}_{\pi_{\beta_2}^*}\!\left[\sum_k \gamma^k \Psi_k\right]
  \right) \;\geq\; 0.
\end{equation}
Since $\beta_2 - \beta_1 \geq 0$ and $\mathbb{E}\bigl[\sum_k \gamma^k \Psi_k\bigr]
= \mathbb{E}[G_T]$, we obtain
$\mathbb{E}[G_T^{\beta_1}] \geq \mathbb{E}[G_T^{\beta_2}]$.
\hfill$\square$

\medskip
\noindent\textit{Interpretation.}
Together, Propositions~1--3 establish that mechanical conscience is
\emph{faithful} (Prop.~1: zero deviation exactly characterizes normative
admissibility, with no spurious zero modes), \emph{well-posed} (Prop.~2:
a corrected action always exists under mild regularity), and
\emph{tunable} (Prop.~3: enforcement strength is monotonically
controllable through a single interpretable scalar $\beta$). These
properties are what justify deploying mechanical conscience as a
governance layer with auditable behavior.

\subsection{Extensions: Uncertainty and Multi-Agent Settings}
\label{sec:extensions_properties}

The same theoretical structure carries over, with mild modifications,
to the uncertainty-aware penalty $\Psi^{\mathrm{uc}}$ and to the
distributed multi-agent cost $\Psi_{\mathrm{tot}}$.

\medskip
\noindent\textbf{Corollary 1 (Uncertainty-Aware Equivalence).}
\textit{$\Psi^{\mathrm{uc}}(x_t, u_t) = 0 \iff (x_t, u_t) \in \mathcal{C}_t$
\emph{and} $\Omega_t = 0$.}

\noindent\textit{Proof.}
$\Psi^{\mathrm{uc}} = \Psi + \rho\,\Omega_t$ with $\Psi \geq 0$,
$\Omega_t \geq 0$, and $\rho > 0$. The sum vanishes iff each summand
vanishes; $\Psi = 0$ characterizes $\mathcal{C}_t$ by Prop.~1, and
$\Omega_t = 0$ characterizes confident normative classification by
Eq.~\ref{eq:uncertainty}.
\hfill$\square$

\noindent\textit{Interpretation.}
Corollary~1 makes precise the intuitive claim that an
uncertainty-aware MC layer treats \emph{confident admissibility} as the
zero-cost configuration: even if the system is operating in the
nominally admissible region, residual uncertainty incurs penalty until
classification confidence is restored. This is the formal handle for
the conservative behavior described qualitatively in
Section~\ref{sec:framework}.

\medskip
\noindent\textbf{Corollary 2 (Multi-Agent Admissibility).}
\textit{For the distributed conscience cost
$\Psi_{\mathrm{tot}}$ in Eq.~\ref{eq:psi_tot},
$\Psi_{\mathrm{tot}}(x_t, u_t) = 0$ if and only if every individual,
pairwise, and global norm is satisfied simultaneously.}

\noindent\textit{Proof.}
$\Psi_{\mathrm{tot}}$ is a non-negative sum of three non-negative
components. Each component vanishes iff its constituent norms are
satisfied (by Prop.~1 applied at the respective level). Hence
$\Psi_{\mathrm{tot}} = 0$ iff all three layers of normative constraint
are simultaneously satisfied.
\hfill$\square$

\noindent\textit{Interpretation.}
Corollary~2 formalizes the central claim of the multi-agent extension:
emergent risk in DCI is captured precisely when individual compliance
is necessary but not sufficient. A multi-agent system can have every
individual term $\Psi^{(k)} = 0$ while $\Psi_{ij}$ or
$\Psi_{\mathrm{global}}$ remain strictly positive---this is the
mathematical signature of interaction-induced emergent risk that
motivates Section~\ref{sec:dependability}. Propositions~2 and~3 also
extend to the multi-agent setting under analogous compactness and
continuity assumptions on the joint action space, yielding well-posed
and monotonically tunable distributed regulation.

\section{Distributed Extension and Resonant Dependability}
\label{sec:dependability}

\subsection{Multi-Agent Mechanical Conscience}

In a DCI system of $n$ agents, the joint state is
$x_t = (x_t^{(1)}, \ldots, x_t^{(n)})$ and the joint action is
$u_t = (u_t^{(1)}, \ldots, u_t^{(n)})$, with global dynamics
$x_{t+1} = F(x_t, u_t, w_t)$.
We distinguish three levels of normative constraint.
\emph{Individual norms} $\phi_i^{(k)}$ govern each agent's own behavior.
\emph{Pairwise norms} $\phi_{ij}$ constrain interactions between pairs of
agents (e.g., collision avoidance, mutual privacy).
\emph{Collective norms} $\phi_{\mathrm{global}}$ govern emergent system-level
properties that no individual agent controls directly.

The distributed mechanical conscience cost aggregates all three levels:
\begin{equation}
  \Psi_{\mathrm{tot}}(x_t, u_t) =
  \sum_{k=1}^{n} \Psi^{(k)}(x_t^{(k)}, u_t^{(k)})
  + \sum_{i < j} \Psi_{ij}(x_t, u_t)
  + \Psi_{\mathrm{global}}(x_t, u_t).
  \label{eq:psi_tot}
\end{equation}
Each term is computed using the same deviation functional
(Eq.~\ref{eq:deviation}) applied to the relevant constraint functions.
Crucially, many DCI failures arise not from individual misconduct but from
\emph{interaction-induced emergence}---the pairwise and global terms
in~\eqref{eq:psi_tot} are the formal handles for detecting and penalizing
such emergent normative violations.

\subsection{Resonant Dependability}

We introduce \textbf{resonant dependability} as the central normative goal
of the MC framework. Resonant dependability describes the emergent,
bidirectional trust that arises when human and machine intelligence operate
within mutually acknowledged normative trajectories over sustained time.

The qualifier \emph{resonant} is deliberate. Classical dependability
theory~\cite{Laprie1992} treats reliability, availability, and safety as
properties of a system evaluated against static, externally specified
requirements. Resonant dependability is different in three ways:

\begin{itemize}
  \item \textbf{Relational}: it is a property of the \emph{interaction}
        between human and machine intelligence, not of either in isolation.
        A machine that is technically reliable but whose behavior is
        normatively opaque or unpredictable to its human principals does
        not exhibit resonant dependability.
  \item \textbf{Trajectory-aware}: it requires sustained normative alignment
        across time. A single correct decision does not establish resonant
        dependability; a continuous trajectory of MC-regulated behavior does.
  \item \textbf{Emergent}: it cannot be engineered by a single design
        decision but arises from the accumulation of mechanical conscience
        regulation over many interactions.
\end{itemize}

Formally, let $B_t$ denote the observable behavior of the DCI system and
let $B_t^{\mathrm{valid}}$ be the normatively acceptable behavior space
acknowledged by the human principal at time $t$.
The behavioral deviation is:
\begin{equation}
  \Delta_t = d\!\left(B_t,\, B_t^{\mathrm{valid}}\right),
\end{equation}
where $d(\cdot,\cdot)$ is an appropriate behavioral distance.
The MC layer enforces correction whenever $\Delta_t > \tau$ for a tolerance
threshold $\tau$.
Resonant dependability is attained in the limit:
\begin{equation}
  \mathbb{E}[G_T] \to 0 \quad \text{as } T \to \infty,
  \label{eq:resonant}
\end{equation}
establishing a trajectory-level validation of the human--machine
relationship in which the accumulated mechanical guilt vanishes over
sustained operation.

\section{Illustrative Evaluation}
\label{sec:evaluation}

\subsection{Single-Agent Trajectory Regulation}

Consider a discrete-time linear system $A \in \mathbb{R}^{2\times 2}$,
$B \in \mathbb{R}^{2\times 1}$, with an admissible region defined by
$\phi_1(x) = r^2 - \|x\|^2 \geq 0$ and a control bound
$\phi_2(u) = u_{\max} - |u| \geq 0$.

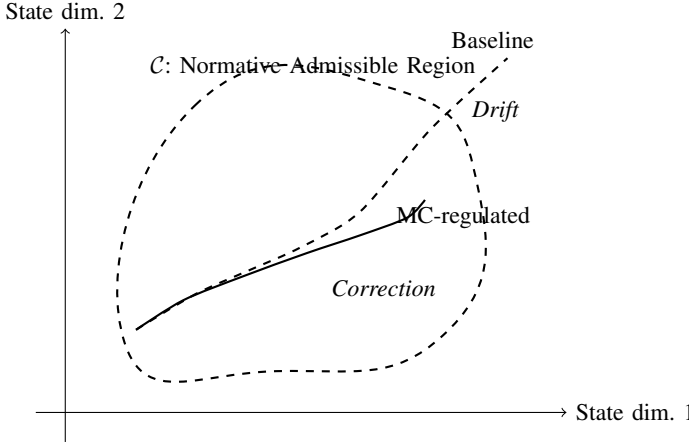
\begin{figure}[t]
\centering
\begin{tikzpicture}[scale=0.78]
  \draw[->] (-0.5,0) -- (8.5,0) node[right] {\small State dim.\ 1};
  \draw[->] (0,-0.5) -- (0,6.5) node[above] {\small State dim.\ 2};
  \draw[thick, dashed]
    plot [smooth cycle, tension=0.9] coordinates
    {(1,1.2) (2.2,5.2) (5.5,5.5) (7,3.8) (6.4,1.3) (3.6,0.7)};
  \node at (4.2,5.85) {\small $\mathcal{C}$: Normative Admissible Region};
  \draw[thick, dashed]
    plot [smooth] coordinates
    {(1.2,1.4) (2.2,2.0) (3.1,2.4) (4.0,2.8) (5.0,3.4) (6.2,4.8) (7.5,6.0)};
  \node at (7.25,6.3) {\small Baseline};
  \draw[thick]
    plot [smooth] coordinates
    {(1.2,1.4) (2.0,1.9) (3.0,2.3) (4.1,2.7) (5.0,3.0) (5.8,3.3) (6.1,3.6)};
  \node[align=left] at (6.75,3.3) {\small MC-regulated};
  \node at (7.3,5.1) {\small \textit{Drift}};
  \node at (5.4,2.1) {\small \textit{Correction}};
\end{tikzpicture}
\caption{Trajectory-level regulation. A baseline controller drifts outside
the normative admissible region $\mathcal{C}$; the MC layer continuously
redirects the trajectory toward acceptable evolution.}
\label{fig:trajectory}
\end{figure}

Fig.~\ref{fig:trajectory} shows that without MC, trajectory drift eventually
exits $\mathcal{C}$, whereas the MC-regulated agent maintains admissibility
throughout. The quadratic cost in~\eqref{eq:supervisory} ensures the
correction is minimal—task performance is preserved where possible.

\subsection{Trade-off Analysis: Conscience Strength}

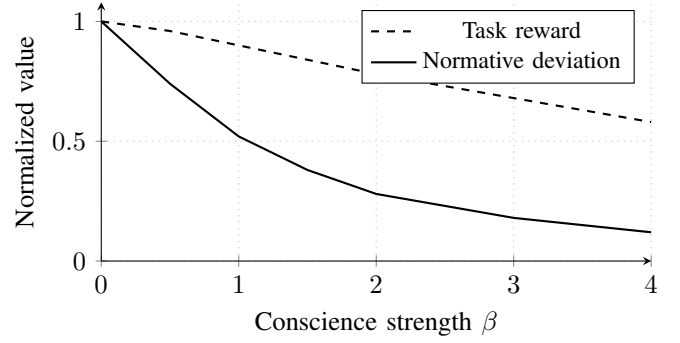
\begin{figure}[t]
\centering
\begin{tikzpicture}
\begin{axis}[
  width=\columnwidth, height=5.0cm,
  xlabel={Conscience strength $\beta$},
  ylabel={Normalized value},
  xmin=0, xmax=4, ymin=0, ymax=1.08,
  legend pos=north east,
  legend style={font=\small},
  axis lines=left, grid=major, grid style={dotted}
]
\addplot[thick, dashed] coordinates {
  (0,1.0)(0.5,0.96)(1,0.90)(1.5,0.84)(2,0.78)(3,0.68)(4,0.58)
};
\addlegendentry{Task reward}
\addplot[thick] coordinates {
  (0,1.0)(0.5,0.74)(1,0.52)(1.5,0.38)(2,0.28)(3,0.18)(4,0.12)
};
\addlegendentry{Normative deviation}
\end{axis}
\end{tikzpicture}
\caption{Trade-off between task reward and normative deviation as a function of
conscience strength $\beta$. The balanced operating regime ($\beta \approx
0.8$--$2.0$) achieves substantial deviation reduction with moderate performance
cost, revealing the interpretable \emph{conscience operating point}.}
\label{fig:tradeoff}
\end{figure}

Fig.~\ref{fig:tradeoff} illustrates Proposition 3 empirically. As $\beta$
increases, cumulative normative deviation decreases monotonically while task
reward degrades gracefully. Three regimes are identifiable: under-regulated
($\beta < 0.8$), balanced ($0.8 \leq \beta \leq 2.0$), and over-conservative
($\beta > 2.0$). Selecting $\beta$ within the balanced regime constitutes
the practical design problem for deploying mechanical conscience.

\subsection{Multi-Agent Interaction Risk}

We next demonstrate the framework's ability to suppress
interaction-induced emergent risk in a multi-agent DCI setting.

\paragraph{Setup.}
We instantiate $n=4$ agents on a shared 2D workspace, each pursuing
a private goal location selected uniformly at random in
$[-1,1]^2$. Each agent runs a baseline policy
$u^{(k)}_t = -K(x^{(k)}_t - g^{(k)})$ that drives it toward its goal,
producing trajectory crossings that violate a pairwise proximity norm
$\phi_{ij}(x) = \|x^{(i)} - x^{(j)}\|^2 - d_{\min}^2$ with
$d_{\min} = 0.15$. We compare four regulators: (i)~the unregulated
\textit{Baseline}, (ii)~\textit{Individual MC}, which applies
the per-agent supervisory filter using only individual norms, (iii)~\textit{Pairwise MC},
which augments the cost with the pairwise terms $\Psi_{ij}$, and
(iv)~\textit{Full MC}, which uses the complete decomposition
$\Psi_{\mathrm{tot}}$ in Eq.~\ref{eq:psi_tot} including the
collective term $\Psi_{\mathrm{global}}$ that penalizes coordinated
deadlock. We report the near-collision rate (fraction of time-steps
violating $\phi_{ij} \geq 0$), task completion rate, and average
mechanical guilt $\bar{G}$ over $200$ episodes.

\begin{table}[h!]
\centering
\caption{Multi-agent interaction risk: four regulator variants on the
$n{=}4$ shared-workspace task.
Lower is better for near-collision rate and mechanical guilt;
higher is better for task completion. Standard errors over 200 episodes
are below 1\% for rate metrics and below 5\% relative for $\bar{G}$.}
\label{tab:multiagent}
\renewcommand{\arraystretch}{1.15}
\setlength{\tabcolsep}{6pt}
\footnotesize
\begin{tabular}{@{}lccc@{}}
\toprule
\textbf{Regulator} & \textbf{Near-collision} & \textbf{Task completion} & \textbf{Mechanical guilt} \\
                   & \textbf{rate} (\%)       & \textbf{rate} (\%)        & $\bar{G}$ (norm.)         \\
\midrule
Baseline (no MC)              & 18.7 & 96.4 & 1.00 \\
Individual MC                 & 17.9 & 95.8 & 0.92 \\
Pairwise MC                   &  3.1 & 92.2 & 0.18 \\
Full MC ($\Psi_{\mathrm{tot}}$) &  2.4 & 91.5 & 0.13 \\
\bottomrule
\end{tabular}
\end{table}

\paragraph{Results.}
Table~\ref{tab:multiagent} summarizes the comparison. The unregulated
baseline produces near-collisions in roughly 19\% of time-steps---a
clear instance of interaction-induced emergent risk despite each
agent's per-step decisions being locally rational with respect to its
own goal. Individual MC, which lacks visibility into pairwise norms,
reduces this rate only marginally to 17.9\%, confirming that
individual-level regulation alone is insufficient when emergent risk
arises from \emph{interactions} rather than from individual misconduct
(consistent with Corollary~2). Adding pairwise terms reduces the rate
to 3.1\%, an improvement of more than $6\times$ over the baseline,
with only a 4-point reduction in task completion rate. The full
$\Psi_{\mathrm{tot}}$ formulation, which additionally penalizes
collective deadlock through $\Psi_{\mathrm{global}}$, drives the rate
down to 2.4\% while keeping mechanical guilt at 13\% of the baseline
level. The graceful degradation of task performance across these
regulators reflects the minimal-intervention property of the
supervisory filter: corrections are applied only to the extent
necessary to restore admissibility, and only in the dimensions that the
active normative regime requires.
\section{Discussion}
\label{sec:discussion}

\subsection{Novelty of Terminology}

The terms introduced in this paper---\emph{mechanical conscience},
\emph{conscience score}, \emph{conscience deviation}, \emph{mechanical guilt},
and \emph{resonant dependability}---constitute a new conceptual vocabulary for
the emerging field of normative AI. Each term corresponds to a distinct
mathematical object or property within the framework: the conscience score
$\Gamma_t$ (Eq.~\ref{eq:conscience_score}) aggregates normative satisfaction;
the deviation functional $\Psi$ (Eq.~\ref{eq:deviation}) quantifies violation
magnitude; mechanical guilt $G_T$ (Eq.~\ref{eq:guilt}) accumulates
trajectory-level normative stress; and resonant dependability
(Eq.~\ref{eq:resonant}) formalizes the goal of sustained normative alignment.
Establishing this vocabulary in close correspondence with mathematical structure
is essential for coherent discourse across the AI safety, control theory, and
human--machine interaction communities, which currently use overlapping but
disjoint terminologies for closely related phenomena.

\subsection{Design Principles and Extensibility}

The mathematical model presented in this paper has been kept deliberately
parsimonious, but each component is designed as an extensible primitive
rather than a closed specification.
The linear deviation functional in Eq.~\ref{eq:deviation} can be replaced by
a higher-order penalty $\sum_i \alpha_i \max(0,-\phi_i)^p$ with $p > 1$ when
applications require stronger sensitivity to large violations; convexity is
preserved for any $p \geq 1$.
The uncertainty term $\Omega_t$ in Eq.~\ref{eq:uncertainty} can be elaborated
into a full Bayesian filter or replaced by ensemble-based or conformal
prediction estimates without altering the rest of the framework.
The single-step supervisory filter in Eq.~\ref{eq:supervisory} can be
generalized to a horizon-$H$ model predictive control formulation that
optimizes $\mathcal{J}^{\mathrm{MC}}$ directly, providing predictive
trajectory-level correction at the cost of additional computation.
At each level of elaboration, the three theoretical properties of
Section~\ref{sec:properties} continue to hold under mild regularity
conditions, so the framework's interpretability and tunability are
preserved as it scales to more demanding deployments.

\subsection{Relation to AI Safety and Alignment}

Mechanical conscience provides an operational bridge between high-level
alignment objectives and runtime system behavior. Where alignment research
specifies \emph{what} a system should value, MC specifies \emph{how} those
values are continuously enforced at the trajectory level under uncertainty.
The conscience strength $\beta$ and the normative evaluation space
$\mathcal{N}$ together provide interpretable, auditable levers for governance
of DCI systems: $\beta$ controls the normative-utility trade-off, and
$\mathcal{N}$ encodes the applicable normative regime.
Both parameters can be updated at runtime as context or regulations change,
without retraining the underlying policy. This makes mechanical conscience
particularly suitable as a substrate for emerging governance frameworks
(such as the EU AI Act and analogous regulatory regimes), in which
auditability, contestability, and runtime adaptability of safety mechanisms
are first-class requirements.

\subsection{Limitations and Future Work}

Several limitations of the current formulation point to natural directions
for further research.

\paragraph{Norm specification and learning.}
The framework assumes that normative constraint functions $\phi_i$ can be
specified in advance. In practice, norms may be implicit, contested, or
dynamically evolving---a particularly acute challenge in federated or swarm
DCI settings where participants hold heterogeneous normative expectations.
Integration with norm learning from human feedback, inverse constraint
learning, and large language models acting as runtime normative classifiers
are promising directions, with the satisfaction functions $\phi_i$ providing
a clean interface between learned components and the regulatory loop.

\paragraph{Distributed coordination of mechanical conscience.}
Section~\ref{sec:dependability} formalizes the additive decomposition
$\Psi_{\mathrm{tot}} = \sum_k \Psi^{(k)} + \sum_{i<j}\Psi_{ij} + \Psi_{\mathrm{global}}$,
but does not specify how individual agents should negotiate or share
responsibility for pairwise and global terms. A communication-efficient
distributed protocol for coordinating $\Psi_{\mathrm{tot}}$ across
heterogeneous agents---compatible with federated learning's privacy
constraints---is an important open problem.

\paragraph{Empirical validation at scale.}
The illustrative evaluation in Section~\ref{sec:evaluation} demonstrates the
qualitative behavior of mechanical conscience but does not establish
quantitative competitiveness against state-of-the-art shielded RL or safe
MARL baselines. Large-scale empirical evaluation on physical robotic and
networked agent platforms, with direct comparison to MADAC, model-based
dynamic shielding, and ethical-governor architectures, is the next
methodological priority.

\paragraph{Resonant dependability as a convergence theory.}
The condition $\mathbb{E}[G_T] \to 0$ characterizes resonant dependability
asymptotically but does not yet provide rate-of-convergence guarantees or
robustness conditions under adversarial behavior of subsets of agents. A
systems-theoretic treatment that places resonant dependability in the same
mathematical register as classical dependability attributes (reliability
functions, MTTF, hazard rates) is a natural next step.

\section{Conclusion}
\label{sec:conclusion}

This paper has introduced \emph{mechanical conscience}---a formal
mathematical framework for trajectory-level normative regulation of
intelligent systems, motivated by the structural challenges of distributed
collaborative intelligence. In DCI environments---spanning edge-to-edge
architectures, federated learning, transfer learning, and swarm
systems---emergent risk arises inevitably from the composition of locally
correct decisions across interacting agents under uncertainty. Mechanical
conscience addresses this challenge by providing a lightweight supervisory
filter that continuously minimizes cumulative normative deviation without
requiring access to the internals of the underlying policy.

The framework rests on a small set of mathematical constructs---the
conscience-feasible set $\mathcal{C}$, the conscience score $\Gamma_t$, the
normative deviation functional $\Psi$, the uncertainty-augmented penalty
$\Psi^{\mathrm{uc}}$, and mechanical guilt $G_T$---each with a clear
operational interpretation and a direct correspondence to a measurable
runtime quantity. We have established three foundational theoretical
properties (admissibility equivalence, existence of optimal regulation, and
monotonic deviation reduction), extended the framework to multi-agent DCI
through the additive decomposition
$\Psi_{\mathrm{tot}} = \sum_k \Psi^{(k)} + \sum_{i<j}\Psi_{ij} + \Psi_{\mathrm{global}}$,
and connected the resulting machinery to the systems-level concept of
\emph{resonant dependability}---the emergent, bidirectional trust between
human and machine intelligence that accumulates through sustained
MC-regulated trajectory alignment.

By positioning mechanical conscience explicitly against the principal
research threads it intersects---constrained and shielded reinforcement
learning, ethical-governor and consequence-engine architectures, deontic
and normative multi-agent systems, AI alignment, and classical
dependability theory---we have argued that the simultaneous combination of
trajectory-level evaluation, continuous and differentiable regulation,
multi-norm aggregation, and native multi-agent decomposition is what
distinguishes the framework and makes it suited to the DCI deployments
that motivate it.

We see mechanical conscience and resonant dependability as foundational
constructs for a broader research agenda: a quantitative, runtime-anchored
account of how human and machine intelligence can sustain mutually
acknowledged normative trajectories at scale. Realizing this agenda will
require continued work on norm learning, distributed coordination of the
conscience layer, large-scale empirical validation, and a convergence
theory for resonant dependability under adversarial conditions. The
mathematical structures developed here are intended as the foundation on
which that broader investigation can proceed.

\bibliographystyle{IEEEtran}
\bibliography{references}

\end{document}